\documentclass[letterpaper, 10 pt, conference]{ieeeconf}  

\usepackage[whole]{bxcjkjatype} 

\IEEEoverridecommandlockouts                              

\usepackage{graphics} 
\usepackage{epsfig} 
\usepackage{mathptmx} 
\usepackage{times} 
\usepackage{amsmath} 
\usepackage{amssymb}  

\usepackage{multirow}
\usepackage[table,xcdraw]{xcolor}
\usepackage{threeparttable}
\usepackage{eucal}
\usepackage{balance}
\usepackage{mathrsfs}  
\usepackage{comment}
\usepackage[noend]{algpseudocode}
\usepackage{float}
\usepackage{textcomp}
\usepackage{mathcomp}
\usepackage{bm}
\usepackage{cite} 
\usepackage{booktabs}
\usepackage{url}

\usepackage{hyperref}

\def\eg{\textit{e.g}\onedot} 
\def\ie{\textit{i.e}\onedot}

\makeatletter
\let\MYcaption\@makecaption
\makeatother

\usepackage[font=footnotesize]{subcaption}

\makeatletter
\let\@makecaption\MYcaption
\makeatother

\usepackage{subcaption}

\algnewcommand{\Break}{\textbf{break}}

\newcommand{\argmin}{\mathop{\rm argmin}\limits}

\newcommand{\myvec}[1]{\mathbf{#1}}
\newcommand{\myarray}[1]{\textbf{\textit{#1}}}
\newcommand{\myset}[1]{\mathcal{#1}}

\def\eg{{\it e.g.}}

\def\ie{{\it i.e.}}

\usepackage[colorinlistoftodos]{todonotes}

\title{\LARGE \bf
Stein Variational Guided Model Predictive Path Integral Control:\\
Proposal and Experiments with Fast Maneuvering Vehicles
}

\author{Kohei Honda$^{1}$, Naoki Akai$^{1}$, Kosuke Suzuki$^{1}$, \\ Mizuho Aoki$^{1}$, Hirotaka Hosogaya$^{1}$, Hiroyuki Okuda$^{1}$, and Tatsuya Suzuki$^{1}$
\thanks{*This work was supported by Tateishi Science and Technology Foundation Research Grants (C) and JSPS KAKENHI Grant Number JP23KJ1067. }
\thanks{$^{1}$The Department of Mechanical Systems Engineering, Graduate School of Engineering, Nagoya University, Furo-cho, Chikusa-ku, Nagoya, Aichi, Japan, {\tt\small honda.kohei.b0@s.mail.nagoya-u.ac.jp}}%
}

\begin{document}

\maketitle

\thispagestyle{empty}
\pagestyle{empty}


\begin{abstract}
This paper presents a novel Stochastic Optimal Control (SOC) method based on Model Predictive Path Integral control (MPPI), named Stein Variational Guided MPPI (SVG-MPPI), designed to handle rapidly shifting multimodal optimal action distributions. While MPPI can find a Gaussian-approximated optimal action distribution in closed form, \ie, without iterative solution updates, it struggles with the multimodality of the optimal distributions. This is due to the less representative nature of the Gaussian. To overcome this limitation, our method aims to identify a target mode of the optimal distribution and guide the solution to converge to fit it. In the proposed method, the target mode is roughly estimated using a modified Stein Variational Gradient Descent (SVGD) method and embedded into the MPPI algorithm to find a closed-form ``mode-seeking'' solution that covers only the target mode, thus preserving the fast convergence property of MPPI. Our simulation and real-world experimental results demonstrate that SVG-MPPI outperforms both the original MPPI and other state-of-the-art sampling-based SOC algorithms in terms of path-tracking and obstacle-avoidance capabilities. \\
\url{https://github.com/kohonda/proj-svg_mppi}



\end{abstract}


\section{INTRODUCTION}
\label{sec:introduction}
Path tracking and obstacle avoidance are essential capabilities required for autonomous mobile robots. These tasks become especially challenging for fast maneuvering vehicles because the optimal action distribution may be multimodal and rapidly shifting. To solve these tasks, sampling-based Model Predictive Control (MPC)~\cite{fox1997dynamic, howard2008state} is a widely adopted approach that can handle the non-linearity and non-differentiability of the environment, such as system dynamics and cost maps, in contrast to gradient-based methods~\cite{diehl2011numerical, andersson2019casadi, CGMRES, honda2023mpc}.

Among various sampling-based MPCs, sampling-based Stochastic Optimal Control (SOC) is a relatively sample-efficient approach that approximates the optimal action distribution as the solution, based on a given prior action distribution~\cite{hansen2003reducing, botev2013cross, williams2018information}. 
In particular, Model Predictive Path Integral control (MPPI)~\cite{williams2018information} stands out as a promising framework because it can estimate a Gaussian-approximated optimal action distribution by analytically minimizing the Kullback-Leibler (KL) divergence. 
That is, it can find the optimal solution in \emph{closed form} (\ie, without iterative solution updates, similar to gradient descent) when given a sufficient number of samples. 

\begin{figure}[t]
  \centering
  \includegraphics[width=0.95\linewidth]{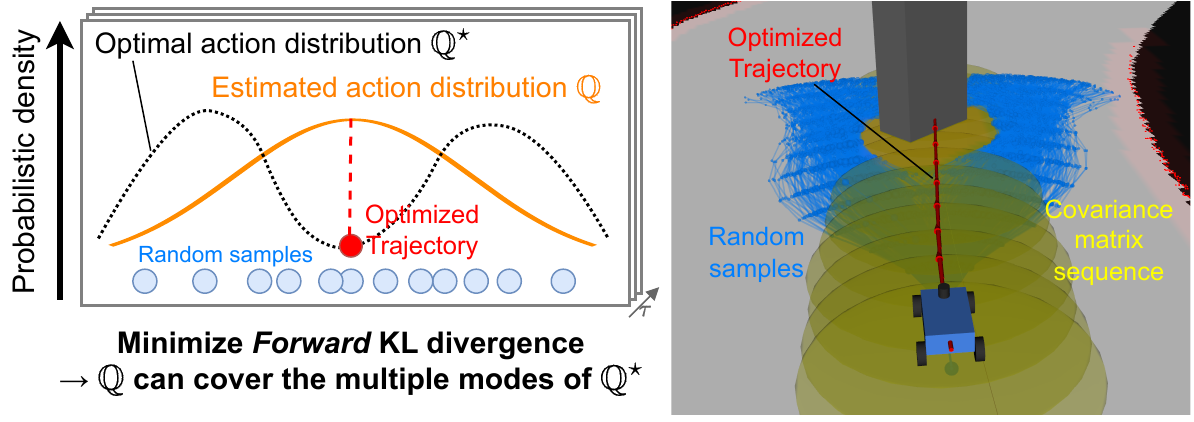}
 \vspace{-3mm}
	\caption{An open issue of MPPI. The MPPI algorithm minimizes the \emph{Forward} KL divergence instead of the original stochastic optimal control problem. As a result, the estimated action distribution $\mathbb{Q}$ may cover the multiple modes of the optimal action distribution $\mathbb{Q^*}$, and it may lead to finding the collision trajectory as the \emph{optimized} one.} 
 \vspace{-3mm}
  \label{fig:mppi_issue}
\end{figure}

\begin{figure}[t]
\centering
  \begin{minipage}[b]{\linewidth}
    \centering
    \includegraphics[width=\linewidth]{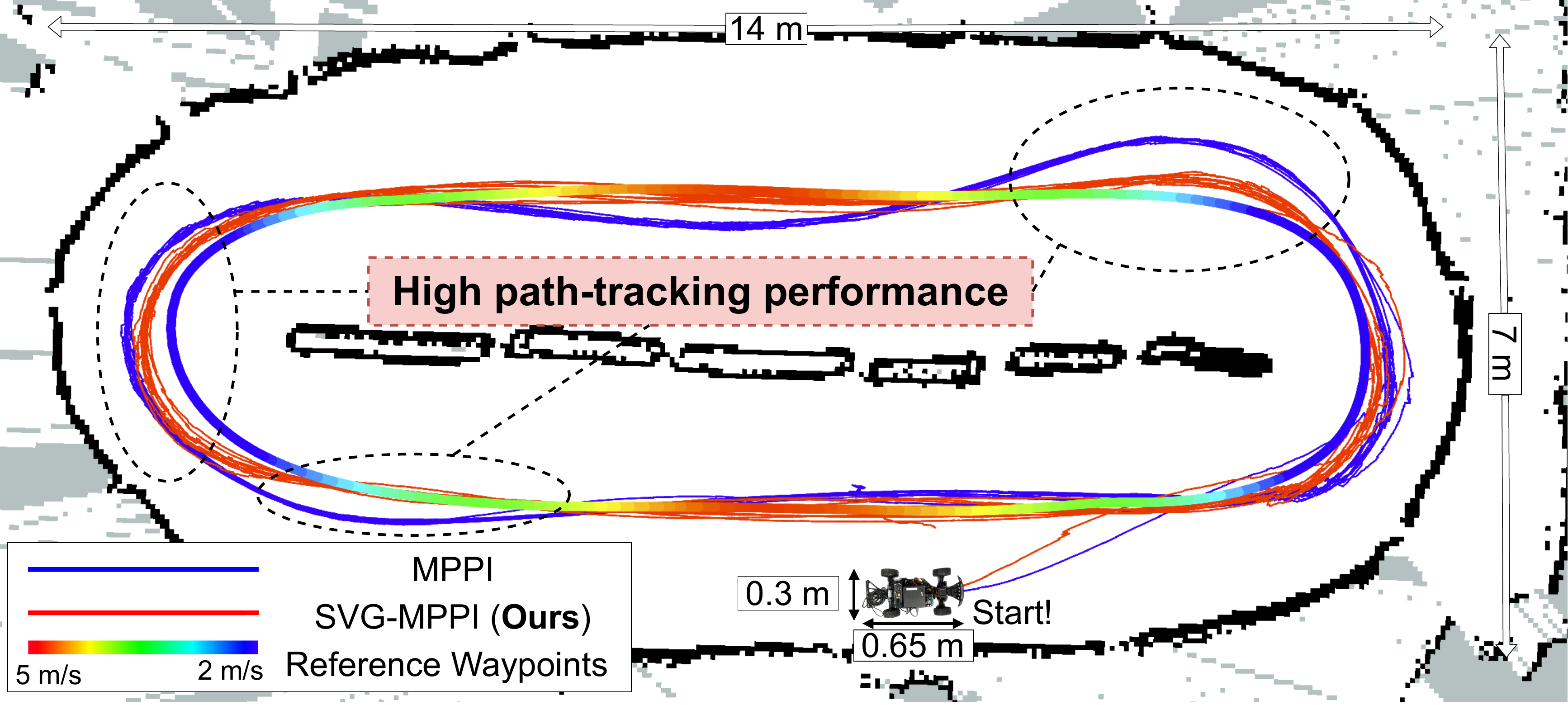}
    \subcaption{Path tracking scenario}
    \label{fig:traj_pt}
  \end{minipage}
  \begin{minipage}[b]{\linewidth}
    \centering
    \includegraphics[width=\linewidth]{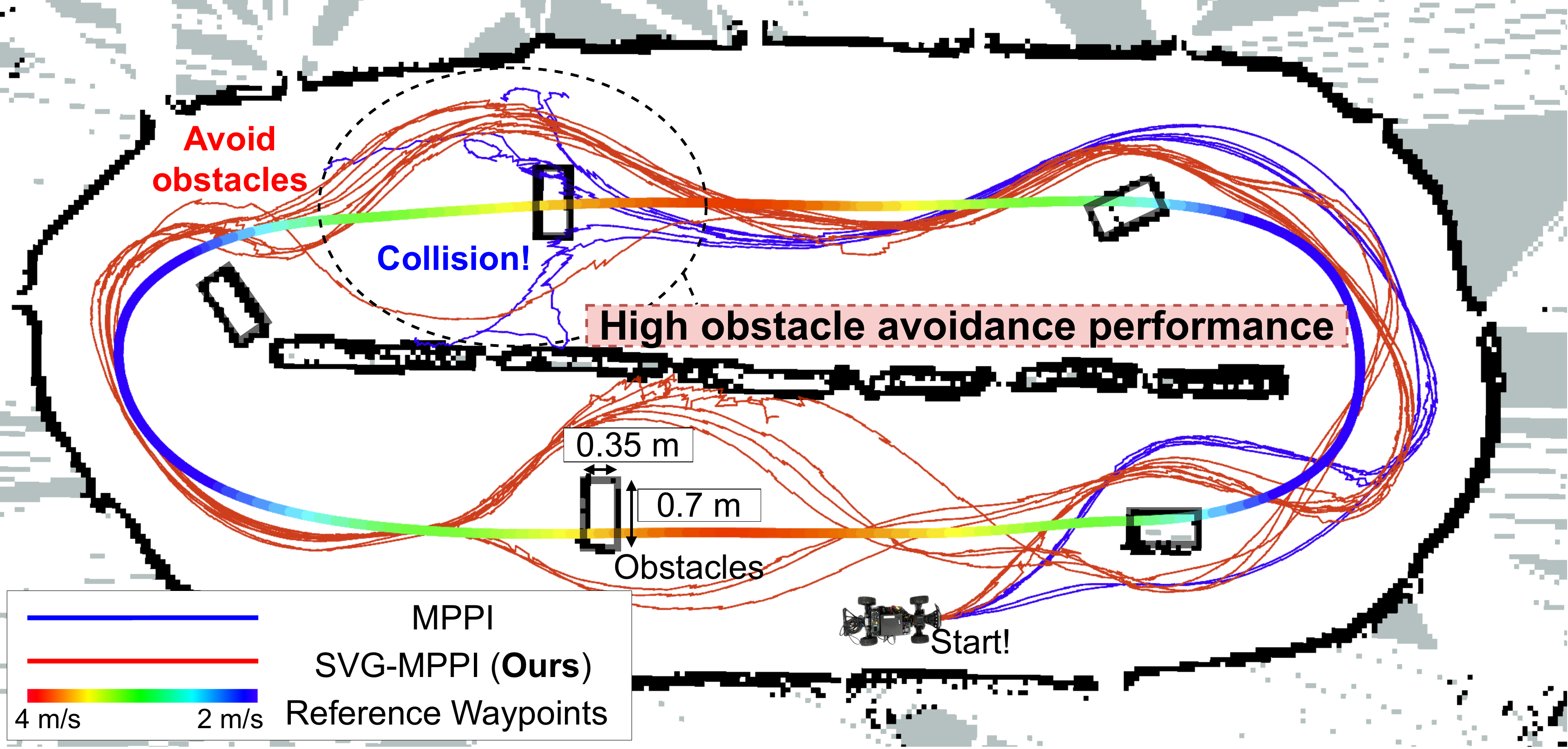}
    \subcaption{Obstacle avoidance scenario}
    \label{fig:traj_oa}
  \end{minipage}
 \vspace{-5mm}
  \caption{Vehicle trajectories in the real-world experiment. Our proposed method, named SVG-MPPI, shows smaller curve overshoots than the original MPPI (Fig.~\ref{fig:traj_pt}) and robustly avoids obstacles where vanilla MPPI collides (Fig.~\ref{fig:traj_oa}). These results can be found at: \href{https://www.youtube.com/watch?v=ML\_aOYQIDL0}{https://www.youtube.com/watch?v=ML\_aOYQIDL0}}
  \label{fig:real_vehicle_traj}
   \vspace{-7mm}
\end{figure}

MPPI, however, has limitations in capturing complex optimal distributions due to its less representative of the Gaussian. For instance, as illustrated in Fig.~\ref{fig:mppi_issue}, an MPPI-based motion planner may produce a collision trajectory as its \emph{optimized} trajectory because the Gaussian-approximated action distribution encompasses the multimodal optimal distribution (see Section~\ref{sec:open_issue} for details). To address these shortcomings, several studies have proposed advanced techniques that capture the multimodality using more representative prior distributions~\cite{lambert2021stein, osa2020multimodal, okada2020variational, wang2021variational, barcelos2021dual} or find \emph{mode-seeking} solutions by leveraging the asymmetry of the KL divergence~\cite{kobayashi2022real}. However, unlike MPPI, these methods cannot find the solution in closed form.
In other words, they require iterative solution updates with sampling and evaluation to converge the solution. 
The iterative updates degrade the convergence of the solution compared to MPPI because the solution is suboptimal and the sequential iteration process restricts the parallelism and the number of samples per iteration. 

In this paper, we propose Stein Variational Guided MPPI (SVG-MPPI), a sampling-based SOC method based on MPPI to address rapidly shifting multimodal optimal action distributions. 
Contrary to the existing methods that capture the complex multimodal distribution, our approach aims to narrow down a single target mode within the multimodal distribution and approximate it with a Gaussian distribution by the MPPI algorithm. 
As a result, our method can obtain a mode-seeking action distribution in closed form, \ie, preserving the fast solution convergence property of MPPI.

Specifically, our proposed method first roughly identifies the target mode by utilizing a small set of samples (hereafter referred to as \emph{guide particles}) and a modified Stein Variational Gradient Descent (SVGD) method~\cite{liu2016stein}. The SVGD method transports the guide particles to near the peak of the target mode within the optimal distribution using its surrogate gradients. Based on the insight that the transport trajectory leading to the peak represents a part of the shape of the target mode, we estimate the rough variance of the target mode using the Gaussian fitting method. Our method then incorporates the peak and variance of the target mode into the MPPI algorithm to converge the solution to cover only the target mode.

In summary, the main contribution of this work is as follows. We propose an MPPI-based SOC method capable of efficiently capturing a single mode of the optimal action distribution. The mode-seeking solution is achieved in closed form by guiding it using a modified SVGD algorithm. Our method has been validated through simulation and real-world experiments, focusing on path tracking and obstacle avoidance tasks for a 1/10th scale vehicle\footnote{Our system was deployed in a public competition called F1TENTH (\url{https://f1tenth.org/}) held at IEEE ICRA 2023, London, where we placed 4th out of 22 teams.}. Empirical results demonstrate that SVG-MPPI outperforms standard MPPI~\cite{williams2018information} and other state-of-the-art SOC algorithms~\cite{lambert2021stein, kobayashi2022real} regarding the path-tracking and obstacle-avoidance performances.
\section{RELATED WORK}
\label{sec:related_work}

To address the rapidly shifting complex optimal action distribution in the sampling-based SOC framework, existing approaches can be broadly categorized into the following three types:


\subsubsection{Increasing Effective Samples}
One strategy for improving the MPPI algorithm involves increasing the number of low-state-cost and feasible samples to find better solutions. Besides learning task-specific prior distributions offline~\cite{power2023variational, sacks2023learning}, the practical approach involves the online adaptation of random samples from a fixed prior distribution, \eg, using auxiliary controllers~\cite{gandhi2021robust, yin2022trajectory, rastgar2024priest}, gradient descent techniques~\cite{bharadhwaj2020model, heetmeyer2023rpgd}, or adaptive importance sampling~\cite{asmar2023model}. However, most methods are limited to differentiable optimal control problems. In contrast, SV-MPC~\cite{lambert2021stein} can transport samples online even in non-differentiable cases by using surrogate gradients. This method, though, leads to a significant increase in computational cost per sample, which limits the number of samples. Instead of transporting all samples, our approach updates a small number of samples (guide particles) and incorporates them into the MPPI algorithm. This approach allows for the exploration of good solutions while suppressing the increase in computational cost. Note that our method is also applicable to non-differentiable problems.


\subsubsection{Approximating Multimodal Distribution}
The second approach is to approximate the complex optimal action distribution with a more representative model, such as Gaussian mixture models, through variational inference methods~\cite{lambert2021stein, osa2020multimodal, okada2020variational, wang2021variational, barcelos2021dual}. Although these approaches excel at capturing multimodality, they cannot find the solution in closed form and necessitate iterative solution updates, unlike MPPI. Moreover, practical control problems require a deterministic solution for the system input rather than a probabilistic multimodal solution.  Extracting a smoothly shifting single deterministic solution from the obtained multimodal solution is far from trivial. Therefore, a \emph{mode-seeking} solution is necessary for practical application to control problems.

\subsubsection{Finding a Mode-Seeking Solution}
The third approach aims to identify one of the modes of the optimal distribution as the target for convergence, leveraging the asymmetry of the KL divergence~\cite{kobayashi2022real}. The KL divergence exhibits an asymmetric property when its arguments are reversed. This approach can obtain a mode-seeking solution by directly minimizing KL divergence reversed from the ones used in the original MPPI. However, minimizing the reverse KL divergence cannot be solved in closed form. Our approach can obtain a mode-seeking solution without sacrificing the closed-form optimality inherent in MPPI by guiding the solution with the reverse KL divergence.

\section{REVIEW OF MPPI}
\label{sec:preliminary}

Our aim is to find a mode-seeking action distribution by improving the MPPI algorithm because MPPI struggles with multimodal optimal distributions due to its inherent nature.
In this section, we first review the MPPI theory described in the original literature~\cite{williams2018information} and then derive the open issue of MPPI to be addressed in this work. Finally, we provide a key property of the KL divergence to address this issue.

\subsection{Review of the MPPI Theory}
\label{sec:mppi_review}

\subsubsection{Problem Formulation}
\label{subsec:problem_formulation}
The MPPI theory models the control target system as a general discrete time dynamical system $\myvec{x}_{\tau+1} = \myarray{F}(\myvec{x}_{\tau}, \myvec{v}_{\tau}), \ \tau \in \{0, \dots, T-1\}$, where $\myvec{x}_{\tau} \in \mathbb{R}^n$ is the predictive state vector at time $\tau$, $\myvec{v}_{\tau} \in \mathbb{R}^m$ is the input vector at time $\tau$, $T$ is the prediction horizon length, and $\myarray{F}$ denotes the general non-linear and non-differentiable state transition function. 
The input vector $\myvec{v}_{\tau}$ is injected with Gaussian noise: $\myvec{v}_{\tau} \sim \myset{N}(\myvec{u}_{\tau}, \Sigma_{\tau})$, as a prior action distribution, where $\myvec{u}_{\tau} \in \mathbb{R}^m$ is the actual input vector and $\Sigma_{\tau} = \mathrm{diag}\{\sigma_{\tau}^0, \dots, \sigma_{\tau}^{m-1}\} \in \mathbb{R}^{m\times m}$ is a given covariance matrix of the Gaussian noise.

MPPI aims to estimate a Gaussian action distribution $\mathbb{Q}$ from randomly generated $K$ input sample sequences, $\myarray{V} = \{\myarray{V}_k\}_{k=0}^{K-1} \in \mathbb{R}^{K \times T \times m}$, where $\myarray{V}_k = \{\myvec{v}_{\tau}\}_{\tau=0}^{T-1} \in \mathbb{R}^{T \times m}$.
The Probability Density Function (PDF) $q(\myarray{V}_k \mid \myarray{U}_t, \boldsymbol{\Sigma})$ corresponding to $\mathbb{Q}$ is analytically expressed as:
\begin{align}
    & q(\myarray{V}_k \mid \cdot ) = Z^{-1} \exp \left(-\frac{1}{2} \sum_{\tau=0}^{T-1}(\myvec{v}_{\tau} - \myvec{u}_{\tau})^\top \Sigma_{\tau}^{-1} (\myvec{v}_{\tau} - \myvec{u}_{\tau}) \right), \label{def_q}
\end{align}
where $Z=\sqrt{(2\pi)^{mT} | \boldsymbol{\Sigma} |}$ is the normalization term, $\myarray{U}_t = \{\myvec{u}_{\tau}\}_{\tau=0}^{T-1} \in \mathbb{R}^{T \times m}$ is the actual control input sequence, and $\boldsymbol{\Sigma} = \mathrm{diag}\{\Sigma_{\tau}\}_{\tau=0}^{T-1} \in \mathbb{R}^{m T \times m T}$ is a given covariance matrix sequence.
To find an optimal control input sequence $\myarray{U}_t^*$, MPPI solves the following SOC problem:
\begin{align}
& \myarray{U}_t^* = \argmin_{\myarray{U}_t \in \myset{U}} \mathbb{E}_{\myarray{V}_k \sim \mathbb{Q}} \left[\phi(\myvec{x}_T) + \sum_{\tau=0}^{T-1} \myset{L}(\myvec{x}_{\tau}, \myvec{u}_{\tau}) \right], \label{original_opt_problem}    
\end{align}
where $\myset{U}$ is the admissible control set~\cite{kerrigan2001robust}, $\myvec{x}_0 = \myvec{x}_t$ is the observed state at time $t$, $\phi$ is the terminal cost function, and $\myset{L}$ is the stage cost function.
Note that the MPPI theory assumes that the stage cost can be divided into state and input costs as, $\myset{L}(\myvec{x}_{\tau}, \myvec{u}_{\tau}) = c(\myvec{x}_{\tau}) + \frac{\lambda}{2} \left(\myvec{u}_{\tau}^\top \Sigma_{\tau}^{-1} \myvec{u}_{\tau} + \beta_t^\top \myvec{u}_{\tau} \right)$, where $\beta_{\tau}$ is a given parameter independent of $\myvec{u}_{\tau}$.
In the following, the $k$th \emph{sequence state cost} is defined as, $S(\myarray{V}_k) = \phi(\myvec{x}_T) + \sum_{\tau=0}^{T-1} c(\myvec{x}_{\tau})$.

Since (\ref{original_opt_problem}) is difficult to solve directly, MPPI instead minimizes the \textit{Forward} KL (FKL) divergence $\mathbb{D}_{\rm{KL}}(\mathbb{Q}^* \parallel \mathbb{Q})$ that the action distribution $\mathbb{Q}$ covers the optimal action distribution $\mathbb{Q}^{*}$, as follows:
\begin{align}
    \myarray{U}_t^* & \simeq \argmin_{\myarray{U}_t \in \myset{U}} \left \{ \mathbb{D}_{\rm{KL}}(\mathbb{Q}^* \parallel \mathbb{Q}) \right \}. \label{min_kld}
\end{align}

\subsubsection{Analytical PDF of the Optimal Action Distribution}
\label{subsec:optimal_distribution}

To minimize the FKL divergence in (\ref{min_kld}), the optimal PDF $q^*(\myarray{V}_k)$ of the $\mathbb{Q}^{*}$ is derived in the following.
First, the \emph{Free-energy} of the system is defined as, $\myset{F}(\myarray{V}_k) = \log \left( \mathbb{E}_{\mathbb{P}} \left[ \exp \left(-\frac{1}{\lambda}S(\myarray{V}_k) \right) \right] \right)$,
where $\lambda$ is a given temperature parameter and $\mathbb{P}$ is a base distribution, which is roughly analogous to a Bayesian prior.
The PDF of the $\mathbb{P}$ is $q(\myarray{V}_k \mid \Tilde{\myarray{U}}, \boldsymbol{\Sigma})$, where $\Tilde{\myarray{U}} = \{\myvec{\Tilde{u}}_{\tau} \}_{\tau=0}^{T-1} \in \mathbb{R}^{T\times m}$ represents a given nominal control input sequence (hereafter referred to as \emph{nominal sequence}). 
The Free-energy is then expanded by Jensen's inequality, the definition of FKL divergence, and (\ref{def_q}) as,
\begin{align}
    & - \lambda \myset{F}(\myarray{V}_k) \leq \mathbb{E}_{\myarray{V}_k \sim \mathbb{Q}} \left[\phi(\myvec{x}_T) + \sum_{\tau=0}^{T-1} \myset{L}(\myvec{x}_{\tau}, \myvec{u}_{\tau}) \right]. \label{what_is_free_energy}
\end{align}
Because the right hand side of (\ref{what_is_free_energy}) is equivalent to the cost function in (\ref{original_opt_problem}), $- \lambda \myset{F}(\myarray{V}_k)$ represents the lower bound of the original SOC problem.
Thus, the optimal PDF $q^*(\myarray{V}_k)$ is derived from the equality condition of (\ref{what_is_free_energy}) as,
\begin{align}
    & q^*(\myarray{V}_k \mid \Tilde{\myarray{U}}, \boldsymbol{\Sigma}) = \eta^{-1} \exp \left(- \frac{1}{\lambda}S(\myarray{V}_k) \right ) q(\myarray{V}_k \mid \Tilde{\myarray{U}}, \boldsymbol{\Sigma} ), \label{q_star}
\end{align}
where $\eta = \int_{\mathbb{R}^{T \times m}} p(\myarray{V}_k \mid \lambda) \exp \left(-\frac{1}{\lambda}S(\myarray{V}_k) \right) \mathrm{d}V_k$ is the normalization term.

\subsubsection{Forward KL Divergence Mimimization}
\label{subsec:minimize_fKLD}
As mentioned in section~\ref{subsec:problem_formulation}, the approach of MPPI is minimizing the FKL divergence in (\ref{min_kld}), which can be transformed by the definition of the FKL divergence and (\ref{def_q}) into the following closed-form expression:
\begin{align}
    \myarray{U}_t^* &= \argmin_{\myarray{U}_t \in \myset{U}} \mathbb{E}_{\myarray{V}_k \sim \mathbb{Q^*}} \left[\frac{1}{2}\sum_{\tau=0}^{T-1}(\myvec{v}_{\tau} - \myvec{u}_{\tau})^\top \Sigma_{\tau}^{-1} (\myvec{v}_{\tau} - \myvec{u}_{\tau}) \right] \nonumber \\
    & = \mathbb{E}_{\myarray{V}_k \sim \mathbb{Q^*}} [\myarray{V}_k]. \label{opt_u}
\end{align}
Thus, the optimal control input sequence $\myarray{U}_t^*$ is obtained by sampling $\myarray{V}_k$ from the optimal action distribution $\mathbb{Q}^*$ and taking its expected value, as in (\ref{opt_u}).
However, since sampling directly from $\mathbb{Q}^*$ is not possible, importance sampling~\cite{kloek1978bayesian} is used with the $q^*(\myarray{V}_k \mid \Tilde{\myarray{U}}, \boldsymbol{\Sigma})$ in (\ref{q_star}), as follows:
\begin{align}
    & \myarray{U}_t^* = \mathbb{E}_{\myarray{V}_k \sim \mathbb{Q}} \left[ \frac{q^*(\myarray{V}_k \mid \Tilde{\myarray{U}}, \boldsymbol{\Sigma})}{q(\myarray{V}_k \mid \hat{\myarray{U}}, \boldsymbol{\Sigma})} \myarray{V}_k \right] \simeq \sum_{k=0}^{K-1} w(\myarray{V}_k) \myarray{V}_k, \label{opt_u_by_is} \\
    & w(\myarray{V}_k) = \eta^{-1} \exp \left( - \frac{1}{\lambda}S(\myarray{V}_k) - \sum_{\tau=0}^{T-1} (\hat{\myvec{u}}_{\tau} - \myvec{\Tilde{u}}_{\tau})^\top \Sigma_{\tau}^{-1} \myvec{v}_{\tau}  \right), \label{weight}
\end{align}
where $\hat{\myarray{U}} = \{\hat{\myvec{u}}_{\tau} \}_{\tau=0}^{T-1} $ is the initial estimated control input sequence, often using the previous optimal solution. 
Importantly, based on the law of large numbers, the MPPI algorithm can obtain a globally optimal solution in the sense of minimizing the FKL divergence in closed form, \ie, the optimal solution can be obtained by sampling sufficiently large $K$ samples once and taking their weighted average, without iterative solution updates.

\subsection{An Open Issue of MPPI}
\label{sec:open_issue}

\begin{figure}[t]
  \centering
  \includegraphics[width=0.8\linewidth]{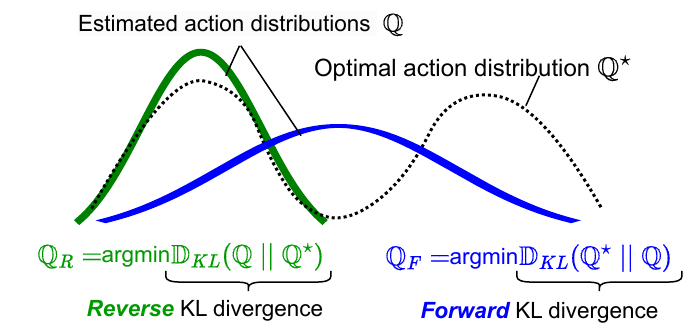}
  \vspace{-4mm}
 \caption{Asymmetry properties of the KL divergence~\cite{kobayashi2022real}. When minimizing the \emph{Forward} KL (FKL) divergence, it results in covering multiple modes within the optimal action distribution. This is because FKL divergence imposes a significant penalty when $q^* > 0$ and $q \approx 0$. In contrast, minimizing the \emph{Reverse} KL (RKL) divergence converges to a single mode in the optimal distribution, a property known as \emph{mode-seeking}. This happens because RKL divergence takes on a larger value when $q > 0$ and $q^* \approx 0$.} 
 \vspace{-6mm}
  \label{fig:kl_divergence}
\end{figure}

The MPPI algorithm can estimate the Gaussian-approximated optimal distribution by analytically minimizing the FKL divergence in closed form. However, the Gaussian approximation causes an open issue that the estimated action distribution may cover the multiple modes of the optimal distribution, as illustrated in Fig.~\ref{fig:kl_divergence}. In the context of an obstacle avoidance task, this issue may lead MPPI to identify the collision trajectory as the optimal solution, as shown in Fig.\ref{fig:mppi_issue}. This occurs because the two branching paths for avoiding obstacles correspond to the peaks of the two modes, and the estimated action distribution covers both modes. To solve this problem, we need to find a mode-seeking action distribution that covers a single target mode of the multimodal distribution.

\begin{figure*}[t]
  \centering
  \includegraphics[width=\linewidth]{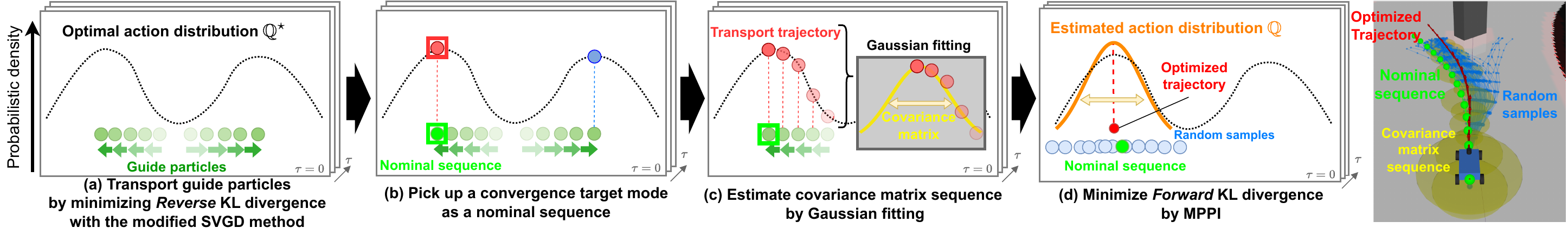}
    \vspace{-6mm}
    \caption{Overview of the proposed method. (a) SVG-MPPI first transports guide particles by minimizing the RKL divergence using the modified SVGD method. These transport trajectories of the guide particles are employed to roughly estimate a convergent target mode. (b) We then identify the peak of the target mode by simply taking the lowest sequence cost guide particle as a nominal sequence. (c) The variances are also roughly estimated by the Gaussian fitting algorithm with the transport trajectory. (d) Finally, SVG-MPPI minimizes the FKL divergence, incorporating the nominal sequence and adaptive covariance matrices in the process. The SVG-MPPI can efficiently find a mode-seeking action distribution compared to the original MPPI (Fig.~\ref{fig:mppi_issue}). }
    \label{fig:method_overview}
    \vspace{-5mm}
\end{figure*}

\subsection{Asymmetry Property of KL divergence}
\label{subsec:RKLD}
One key to finding the mode-seeking solution is to utilize the asymmetry property of the KL divergence. 
The standard MPPI algorithm minimizes \emph{Forward} KL (FKL) divergence, denoted as $\mathbb{D}_{\rm{KL}}(\mathbb{Q}^* \parallel \mathbb{Q})$, where $\mathbb{Q}$ may cover the entire $\mathbb{Q}^*$. 
In contrast, when the input arguments are reversed, the \emph{Reverse} KL (RKL) divergence $\mathbb{D}_{\rm{KL}}(\mathbb{Q} \parallel \mathbb{Q}^*)$ exhibits a different characteristic, where $\mathbb{Q}$ converges to a single mode of the optimal distribution $\mathbb{Q}^*$, as shown in Fig.~\ref{fig:kl_divergence}.
\section{STEIN VARIATIONAL GUIDED MPPI}
\label{sec:method}

Although some existing methods~\cite{lambert2021stein, kobayashi2022real} can find the mode-seeking solution by directly minimizing the RKL divergence described in Section~\ref{subsec:RKLD}, the solution cannot be obtained in closed form, resulting in a degradation of convergence.
To overcome the limitation, our approach guides the MPPI solution to converge to a single target mode roughly identified using a modified SVGD method~\cite{liu2016stein}.

As an overview of our method illustrated in Fig.~\ref{fig:method_overview}, we first spread a small set of samples (guide particles) and transport them by the modified SVGD method. Because the SVGD method iteratively reduces the RKL divergence based on the surrogate gradients of the optimal action distribution $\mathbb{Q}^*$, one of the guide particles is transported to near the peak of the target mode. We then roughly estimate the variance of the target mode by fitting the transport trajectory leading to the peak as Gaussian. Finally, we incorporate the peak and variance of the target mode (as the nominal sequence and covariance matrix sequence) into the MPPI algorithm to converge the solution to cover only the target mode.

\subsection{Transport Guide Particles by the Modified SVGD method}
\label{subsec:transport}
Our proposed method first spreads $K_g$ guide particles $\{\myarray{V}_k^g\}_{k=0}^{K_g-1} \in \mathbb{R}^{K_g \times T \times m}$ in addition to the $K$ samples used in the standard MPPI algorithm.
Note that $K_g \ll K$, in particular, we use $K_g=1$ in our experiments in section \ref{sec:experiment}.
The guide particles are transported to make the RKL divergence smaller by applying the following gradient descent update with $L$ iterations per control cycle given a small step size $\epsilon$,
\begin{align}
    & \hat{\myarray{V}_k^g} \leftarrow \myarray{V}_k^g - \epsilon \nabla_{\myarray{V}_k^g} \mathbb{D}_{\rm{KL}}(\mathbb{Q} \parallel \mathbb{Q}^*), \; ^\forall k \in \{ 0, \dots , K_g-1\}. \label{transport_samples}
\end{align}
The RKL divergence $\mathbb{D}_{\rm{KL}} \left(\mathbb{Q} \parallel \mathbb{Q}^* \right)$ is expanded from its definition and Jensen's inequality as,
\begin{align}
    \mathbb{D}_{\rm{KL}} \left(\mathbb{Q} \parallel \mathbb{Q}^* \right) 
    &= -\mathbb{E}_{\myarray{V}_k^g \sim \mathbb{Q}} \left[ \log \frac{q^*}{q} \right]
    = - \mathbb{E}_{\myarray{V}_k^g \sim \mathbb{Q}} \left [\log w(\myarray{V}_k^g) \right] \nonumber \\
    &\leq - \log \mathbb{E}_{\myarray{V}_k^g \sim \mathbb{Q}} \left[w(\myarray{V}_k^g) \right], \label{upper_RKLD}
\end{align}
where $w(\cdot)$ is the weight function in (\ref{weight}).
Now, we can reduce the RKL divergence of the guide particles by assigning the gradient of (\ref{upper_RKLD}) into (\ref{transport_samples}) since the right hand side of (\ref{upper_RKLD}) is the upper bound of the RKL divergence.
However, the weight function $w(\cdot)$ has the sequence state cost $S(\myarray{V}_k^g)$, which is generally assumed to be non-differentiable with $\myarray{V}_k^g$.
Therefore, we estimate the stochastic surrogate gradient described in \cite{lambert2021stein}:
\begin{align}
    \nabla_{\myarray{V}_k^g} \mathbb{D}_{\rm{KL}}(\mathbb{Q} \parallel \mathbb{Q}^*) 
    &= - \nabla_{\myarray{V}_k^g}\log \mathbb{E}_{\myarray{V}_k^g \sim \mathbb{Q}} \left[w(\myarray{V}_k^g) \right] \nonumber\\
    &= - \frac{\mathbb{E}_{\myarray{V}_k^g \sim \mathbb{Q}} \left [w \left(\myarray{V}_k^g \right) \nabla_{\myarray{V}_k^g} \log q \right ]}{\mathbb{E}_{\myarray{V}_k^g \sim \mathbb{Q}} \left[w(\myarray{V}_k^g) \right]} \nonumber \\
    & \simeq - \frac{\sum_{i=0}^{N-1} \left\{ w \left(\myarray{V}_k^g[i] \right) \nabla_{\myarray{V}_k^g} \log q \left(\myarray{V}_k^g[i] \right) \right \}}{\sum_{i=0}^{N-1} w\left(\myarray{V}_k^g[i] \right) }, \label{surrogate_grad} 
\end{align}
where $\myarray{V}_k^g[i]$ is a sample for the Monte Carlo estimation, \ie, $\myarray{V}_k^g[i] \sim q \left(\myarray{V}_k^g[i] \mid \myarray{V}_k^g, \Sigma_g \right)$, given a constant covariance matrix $\Sigma_g$.
The update equation in (\ref{transport_samples}) converges to a simplified version of the update equation using the SVGD method in \cite{lambert2021stein}, with the kernel function and log-prior terms removed.
These removed terms penalize particle cohesion and preserve the diversity of the samples. 
However, for our approach, the penalty terms are arbitrarily removed because we expect the guide particles to aggregate to some of the peaks of the optimal distribution.
Note that the computational complexity for the $L$ times transport in (\ref{transport_samples}) is $\mathcal{O}(L K_g N)$ and relatively tiny compared to the MPPI's one (\ie, $\mathcal{O}(T K m)$) when $K_g \ll K$.

\subsection{FKL divergence Minimization with the Nominal Sequence and Adaptive Covariance Matrix Sequence}

We now obtain $K_g$ transport trajectories for each guide sample as, $\tau^g_k = \left\{ \myarray{V}_k^g[l] \right\}_{l=0}^{L-1} \in \mathbb{R}^{L \times T \times m}, k \in \{0, \dots, K_g-1\}$, where $L$ is the number of iterations for (\ref{transport_samples}).
We then roughly identify the peak and variance of the target mode using the transport trajectories and incorporate them as the nominal sequence $\Tilde{\myarray{U}}$ and covariance matrix sequence $\boldsymbol{\Sigma}$ in the MPPI algorithm described in Section~\ref{sec:mppi_review}. 

\subsubsection{Picking a Target Mode}
The guide particles transported in sufficiently large iterations $L$ are expected to be close to some of the peaks. Since the convergence goal of a gradient descent depends on the initial value, the guide particles can be transported to different peaks within the optimal distribution after the iterations. 
The strategy for determining the peak of the target mode is simply to pick up a transported guide particle that has the lowest sequence state cost. 
Therefore, the peak, as the nominal sequence $\Tilde{\myarray{U}}$ of the MPPI algorithm, is determined, as follows:
\begin{align}
    & \Tilde{\myarray{U}} = \myarray{V}^g_{k^*}, \text{ where }  k^* = \argmin_{k}{ \left \{S(\myarray{V}_k^g[L-1]) \right \}_{k=0}^{K_g-1}}. \label{estimated_nominal_solution}
\end{align}
The nominal sequence $\Tilde{\myarray{U}}$ is used in the weight function (\ref{weight}), which penalizes the deviation of the nominal sequence (the peak of the target mode) and the mean of the action distribution, which corresponds to determining the coverage target mode within the optimal action distribution.

\subsubsection{Estimation of the Adaptive Covariance Matrix Sequence}
To converge the action distribution to cover only the target mode, we should set an appropriate covariance matrix sequence $\boldsymbol{\Sigma}$, which is used in the prior action distribution of random samples and weight function in (\ref{weight}) for the MPPI algorithm. Our method utilizes the entire $k^*$th transport trajectory $\tau^g_{k^*}$ to estimate the rough variance of the target mode because the transport trajectory leading to the peak is expected to represent a part of the shape of the target mode.
Specifically, we use a Gaussian fitting algorithm known as fast and rough~\cite{guo2011simple}.
For all predictive times and control inputs $^\forall \tau \in [0, \dots, T-1]$ and $^\forall i \in [0, \dots, m-1]$,
let the input vector be $\myvec{a} = \left[\myarray{V}_{k^*}^g[0, \tau, i], \dots, \myarray{V}_{k^*}^g[L-1, \tau, i] \right] \in \mathbb{R}^L$ and the corresponding $q^*$ in (\ref{q_star}) be $\myvec{b} = \left[q^*\left(\myarray{V}_{k^*}^g[0]\right), \dots, q^* \left (\myarray{V}_{k^*}^g[L-1] \right) \right] \in \mathbb{R}^L$, and the variance $\Sigma_{\tau}[i] = \sigma_{\tau}^i$ for each control input can be estimated by solving the first-order equation:
\begin{align} 
& \sigma_{\tau}^i = \sqrt{\frac{-1}{2z_2}}, \text{ where }\label{estimated_cov}\\
& \left[\begin{array}{ccc}
\sum \myvec{b}^2 & \sum \myvec{a} \myvec{b}^2 & \sum \myvec{a}^2 \myvec{b}^2 \\
\sum \myvec{a} \myvec{b}^2 & \sum \myvec{a}^2 \myvec{b}^2 & \sum \myvec{a}^3 \myvec{b}^2 \\
\sum \myvec{a}^2 \myvec{b}^2 & \sum \myvec{a}^3 \myvec{b}^2 & \sum \myvec{a}^4 \myvec{b}^2
\end{array}\right]
\left[\begin{array}{l}
z_0 \\
z_1 \\
z_2
\end{array}\right] 
=\left[\begin{array}{l}
\sum \myvec{b}^2 \log \myvec{b} \\
\sum \myvec{a} \myvec{b}^2 \log \myvec{b} \\
\sum \myvec{a}^2 \myvec{b}^2 \log \myvec{b}
\end{array}\right]. \nonumber 
\end{align}
As a result, when the change of the $q^*$ due to the transport is small, the variance is estimated to be small, as shown in Fig.~\ref{fig:method_overview}. In this case, the sequence state cost of the guide particle is largely shifted by the transport. In the opposite case, the variance is kept to be large.
\section{EXPERIMENTS}
\label{sec:experiment}

\subsection{Experiment Setup}

We implemented our method for a path-tracking and obstacle-avoidance motion planner of a 1/10th-scale vehicle.
The motion planner controls the vehicle's steering angle based on given reference waypoints (path and speed profile), the vehicle's position and velocity, and a cost map~\cite{fankhauser2016universal} constructed from onboard 2D LiDAR scans for obstacles and course detection.

\subsubsection{MPC Formulation}
The implemented vehicle dynamics predictive model $\myarray{F}$ was the kinematic bicycle model~\cite{vehicle_model_Rajamani} considering the dead time and first-order delays with the steering angle input, which have non-lineality.
The non-differentiable sequence state cost was $S(\myarray{V}_k) = \sum_{\tau=0}^{T=15} \left\{\Delta d_r(\tau)^2 + 0.01 \Delta \theta_r(\tau)^2 + 1000 \; \myvec{1}^{\rm{collide}}(\myvec{x}_{\tau}) \right\}$, where $\Delta d_r$ and $\Delta \theta_r$ were position and yaw angle deviations from the reference waypoints, and $\myvec{1}^{\rm{collide}}$ is a binary indicator function for collision, provided by the cost map. The step width of the prediction horizon $\Delta \tau$ is 0.05 s.
Please refer to the source code\footnote{\url{https://github.com/kohonda/proj-svg_mppi}} for the detailed algorithm and settings.

\subsubsection{Validation Scenarios}
To validate the performance of the proposed method, we conducted tests under two types of driving scenarios, both in simulation and in the real world:
\begin{itemize}
\item Path Tracking (PT): The vehicle completed $N_{\rm{pt}}$ laps of the course without encountering unforeseen obstacles.
\item Obstacle Avoidance (OA): The vehicle drove $N_{\rm{oa}}$ laps of the course, encountering five unforeseen static obstacles. These obstacles were randomly repositioned around reference waypoints on each lap.
\end{itemize}

\subsubsection{Evaluation Metrics}
To evaluate the path-tracking and obstacle-avoidance capabilities, we employ the following two evaluation metrics:
\begin{itemize}
\item Mean Sequence state cost $S(\myarray{U}_t)$ per lap, denoted as MS: A lower value of this metric indicates better performance in path tracking and obstacle avoidance. In scenario PT, this metric is primarily focused on evaluating path-tracking performance. Conversely, scenario OA, mainly evaluates the obstacle-avoidance performance.
\item Collision Rate (CR): This represents the rate of collisions with obstacles or the course, relative to the total number of obstacles encountered in scenario OA.
\end{itemize}

\subsubsection{Baseline Methods}
We implemented the vanilla MPPI~\cite{williams2018information}, Reverse-MPPI with sample rejection~\cite{kobayashi2022real}, and SV-MPC~\cite{lambert2021stein} to compare their performance to our proposed method, SVG-MPPI. 
Reverse-MPPI minimizes the RKL divergence based on mirror descent algorithm~\cite{beck2003mirror}, and SV-MPC also directly minimizes RKL divergence with variational inference approach, \ie, using SVGD method~\cite{liu2016stein}. 
Unlike the MPPI that minimizes the FKL divergence, they cannot be solved in closed form, and the number of random samples for the iterative updates must be set smaller than the MPPI and SVG-MPPI to keep real-time processing.

All algorithms, including SVG-MPPI, were developed in ROS and C++ and employed CPU multi-threading via OpenMP~\cite{chandra2001parallel}. The same MPC formulation and common parameters were used across all algorithms, except the number of random samples. To keep a calculation time of approximately 20 ms on a desktop PC equipped with an Intel Core i9-10850K CPU, in the simulation, we set the number of random samples for MPPI, Reverse-MPPI, SV-MPC, and SVG-MPPI at 10k, 200, 500, and 8k, respectively. Similarly, in the real-world experiment, 5k and 2k random samples are used for MPPI and SVG-MPPI, respectively.

\subsection{Simulation Results}
We executed each method for 100 laps in both the PT and OA scenarios to provide a comprehensive comparison ($N_{\rm{pt}} = N_{\rm{oa}} = 100$). For the simulation environment, we customized the F1TENTH Gym environment~\cite{o2020f1tenth} to include unforeseen five obstacles that were randomly replaced around reference waypoints (within 0.1 m) on each lap.

\subsubsection{Comparison with Baseline Methods}
\begin{figure}[t]
  \centering
  \includegraphics[width=1.0\linewidth]{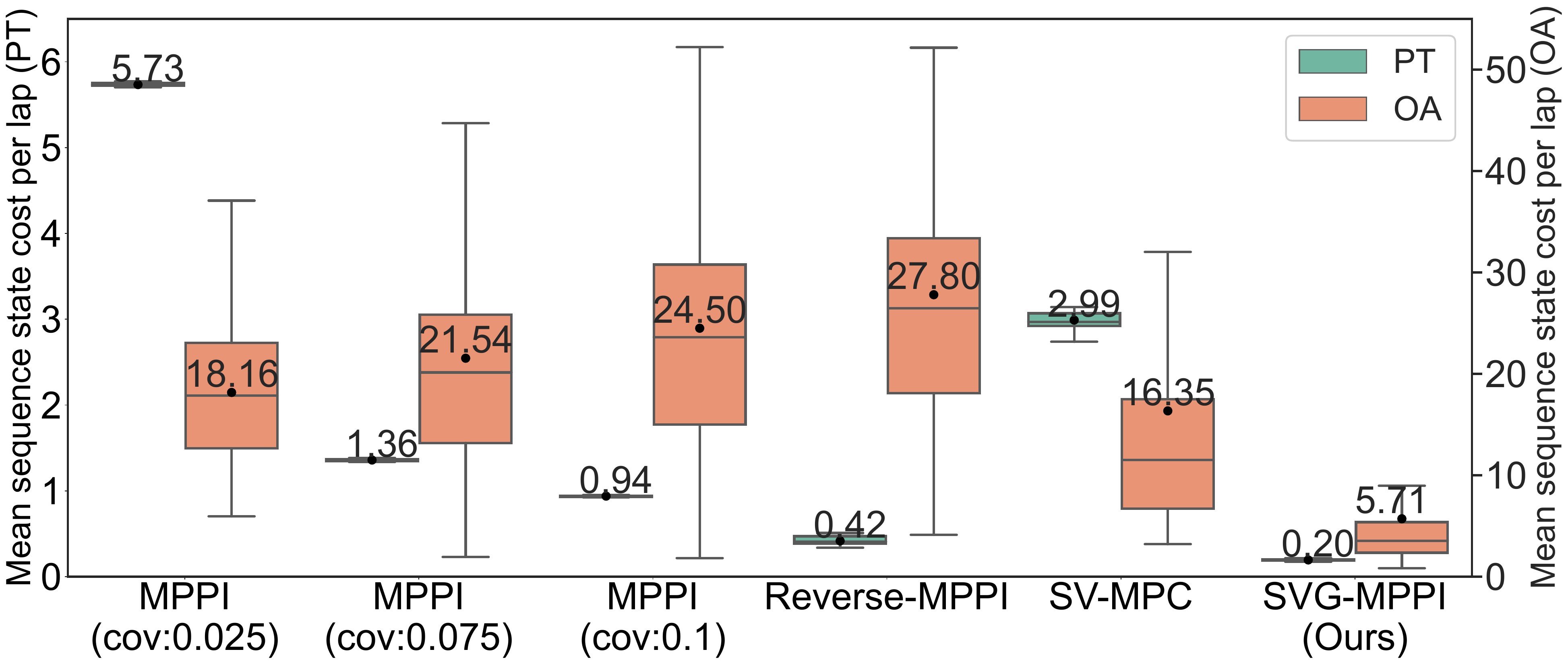}
  \caption{Comparison of the MSs with baseline methods in the simulation.} 
  \label{fig:comparison_with_baselines}
\end{figure}

Figure~\ref{fig:comparison_with_baselines} shows the evaluation results for the MS metrics in both PT and OA scenarios, and Table~\ref{tab:collision_rate} lists the CRs for each method in the OA scenario. In the case of vanilla MPPI, a lower covariance of the control input correlates with lower MSs and CRs in the OA scenario; however, it negatively impacts MSs in PT. These findings suggest that vanilla MPPI faces a trade-off between PT and OA capability depending on the fixed covariance parameter. This is attributable to the fact that a small covariance prevents the solution from covering the entire optimal action distribution in the OA, but is not suitable for PT, which requires covering the entire single-mode distribution to track sharp curves rapidly.

Unlike vanilla MPPI, both Reverse-MPPI and SV-MPC aim to minimize RKL divergence directly. As shown in Fig.~\ref{fig:comparison_with_baselines}, Reverse-MPPI outperforms vanilla MPPI in PT, while SV-MPC outperforms in the OA scenario. However, each method underperforms in the other scenario; Reverse-MPPI in OA and SV-MPC in PT. 
In particular, SV-MPC is similar to our method. It transports all samples and takes a weighted average of them, resulting in them being evenly separated, which can degrade the performance of OA, or the performance of PT can be degraded due to the samples being over-compressed.

On the other hand, our proposed method, SVG-MPPI, demonstrates superior performance in both scenarios, with average MS values of 0.20 and 5.71 for PT and OA, respectively as shown in Fig.~\ref{fig:comparison_with_baselines}. These results confirm that SVG-MPPI excels in both PT and OA tasks, avoiding the trade-offs observed in all baseline methods. 
Furthermore, SVG-MPPI records the lowest CR of 4.0 \% across all trials in the OA scenario in Table~\ref{tab:collision_rate}. In the rare cases where collisions occurred, SVG-MPPI did not find a mode-seeking solution due to its inability to guarantee one.

We also measured the computation time for each method. The average/maximum time to find a solution for each method was 12.1 ms / 34.2 ms for MPPI, 26.0 ms / 55.9 ms for Reverse MPPI, 29.1 ms / 55.9 ms for SV-MPC, and 11.0 ms / 39.7 ms for SVG-MPPI. MPPI and SVG-MPPI can find solutions through closed-form expressions, so even with a large number of samples, the computation time is kept relatively low thanks to the benefits of parallelization. On the other hand, Reverse MPPI and SV-MPC need to update their solutions iteratively, which tends to increase the computational load relative to the number of samples compared to the others.

\begin{table}[t] 
\caption{Collision Rate (CR) for Each Method in the Simulation}
\vspace{-3mm}
\label{tab:collision_rate}
\centering
\begin{tabular}{ccccccc}
\toprule
Method & \multicolumn{3}{c}{MPPI} & \multirow{2}{*}{\begin{tabular}[c]{@{}c@{}}Reverse\\ MPPI\end{tabular}} & \multirow{2}{*}{\begin{tabular}[c]{@{}c@{}}SV\\ MPC\end{tabular}} & \multirow{2}{*}{\begin{tabular}[c]{@{}c@{}}SVG-MPPI\\ (Ours)\end{tabular}} \\
Cov. {[}rad{]} & 0.025 & 0.075 & 0.1 &  &  &  \\
\midrule
CR {[}\%{]} $\downarrow$  & 13.6 & 21.0 & 24.4 & 24.8 & 12.4 & \textbf{4.0} \\ 
\bottomrule
\end{tabular}
\vspace{-3mm}
\end{table}

\subsubsection{Ablation Study}

SVG-MPPI is an extension of vanilla MPPI that incorporates both a Nominal Sequence (NS) and Adaptive Covariance matrix sequence estimation (AC). To demonstrate the complementary nature of these two extensions, we conducted an ablation study for the PT and OA scenarios in the simulation. The results are presented in Table~\ref{tab:ablation_study}. Both MPPI and MPPI+NS employed a fixed steering covariance of 0.75 rad. As indicated in Table~\ref{tab:ablation_study}, MPPI+NS outperforms vanilla MPPI in both PT and OA scenarios. Moreover, MPPI+AC exhibits even better performance in OA compared to MPPI+NS, although its performance in PT is significantly diminished. When both enhancements (MPPI+NS+AC) are incorporated, SVG-MPPI delivers the best performance in OA and ranks second-best in PT. Overall, these results confirm that the NS and AC functions are complementary and contribute to the enhanced path-tracking and obstacle-avoidance performance. However, as the results for SVG-MPPI and MPPI+NS show, AC slightly hinders PT performance. This is because the proposed AC method, prioritizing computational efficiency, is a rough estimation prone to errors. This is a future work to be addressed.

\begin{table}[t]
\caption{Ablation Study: Average of MSs for 100 trials}
\vspace{-3mm}
\label{tab:ablation_study}
\centering
\begin{tabular}{ccc}
\toprule
method & Scenario PT & Scenario OA \\
\midrule
MPPI (baseline) & 1.36 & 21.54 \\
MPPI + nominal sequence (NS) & \textbf{0.18} & 11.03 \\
MPPI + adaptive covariance (AC) & 18.18 & 7.08 \\
MPPI + NS + AC (SVG-MPPI) & 0.20 & \textbf{5.71}\\
\bottomrule
\end{tabular}
\end{table}

\subsection{Real-world Experiment Results}
We also conducted real-world experiments using a 1/10th-scale vehicle equipped with a 2D LiDAR (HOKUYO UTM30-LX) and a compact computer (Intel NUC with a Core i7-1360P Processor). To estimate the vehicle's pose on the pre-built map, we employed an advanced Monte Carlo localization method~\cite{akai2023reliable}.

Figure~\ref{fig:real_vehicle_traj} shows the trajectories of both vanilla MPPI (steering covariance: 0.075 rad) and SVG-MPPI in PT and OA scenarios with a fixed obstacle layout. The figure reveals that SVG-MPPI exhibits smaller overshoots on curves compared to vanilla MPPI, as further illustrated in Fig.~\ref{fig:traj_pt}. Additionally, Fig.\ref{fig:traj_oa} indicates that SVG-MPPI successfully avoids obstacles where vanilla MPPI encounters collisions.

For more evaluation, we completed 10 laps in scenario PT ($N_{\rm{pt}}=10$) and 3 laps for each of six different static obstacle layouts in scenario OA ($N_{\rm{pt}}=18$). In scenario PT, the mean MS values were recorded at 5.83 and 2.03 for vanilla MPPI and SVG-MPPI, respectively. Turning our attention to scenario OA, the mean MS and CR values were (62.73, 55.6 \%) for vanilla MPPI and (27.82, 15.4 \%) for SVG-MPPI. Overall, SVG-MPPI demonstrated superior path-tracking and obstacle-avoidance capabilities compared to the vanilla MPPI method in real-world testing.

\section{CONCLUSION AND POTENTIAL LIMITATION}

This paper presents a novel MPPI-based SOC method to address rapidly shifting multimodal optimal action distributions. Our method is capable of finding a mode-seeking solution in closed form by guiding the convergence target of the MPPI solution using the SVGD method. While our experiments do not show any performance degradation of the proposed method, a potential limitation of our method arises in cases where the gradients of the optimal distribution are zero outside the peaks of the modes. This limitation is due to the fact that the SVGD method primarily tracks gradients, and the solution may become trapped at such terrace-like places within the optimal distribution. Furthermore, since the proposed method is not theoretically tailored for path tracking and obstacle avoidance, it is expected to be applicable to a wider range of robotic challenges.







\balance


\bibliographystyle{IEEEtran}
\bibliography{IEEEabrv, reference}

\end{document}